%% file: main.tex
\documentclass[10pt,twocolumn,letterpaper]{article}

\usepackage[pagenumbers]{wacv} 

\usepackage{graphicx}
\usepackage{amsmath}
\usepackage{amssymb}
\usepackage{booktabs}
\usepackage{tabularx}
\usepackage{multirow}
\newcolumntype{L}[1]{>{\raggedright\let\newline\\\arraybackslash\hspace{0pt}}m{#1}}
\newcolumntype{C}[1]{>{\centering\let\newline\\\arraybackslash\hspace{0pt}}m{#1}}
\newcolumntype{R}[1]{>{\raggedleft\let\newline\\\arraybackslash\hspace{0pt}}m{#1}}

\usepackage{siunitx}
\usepackage[accsupp]{axessibility}

\usepackage[pagebackref,breaklinks,colorlinks]{hyperref}

\usepackage[capitalize]{cleveref}
\crefname{section}{Sec.}{Secs.}
\Crefname{section}{Section}{Sections}
\Crefname{table}{Table}{Tables}
\crefname{table}{Tab.}{Tabs.}

\def\wacvPaperID{590} 
\def\confName{WACV}
\def\confYear{2025}

\begin{document}

\title{ReMP: Reusable Motion Prior for Multi-domain\\3D Human Pose Estimation and Motion Inbetweening}

\author{Hojun Jang$^{1}$ and Young Min Kim$^{1,2}$ \\
\\
$^1$ Dept. of Electrical and Computer Engineering, Seoul National University\\
$^2$ Interdisciplinary Program in Artificial Intelligence and INMC, Seoul National University\\
{\tt\small \{j12040208, youngmin.kim\}@snu.ac.kr}
}
\maketitle

\begin{abstract}
We present Reusable Motion prior (ReMP), an effective motion prior that can accurately track the temporal evolution of motion in various downstream tasks.
Inspired by the success of foundation models, we argue that a robust spatio-temporal motion prior can encapsulate underlying 3D dynamics applicable to various sensor modalities.
We learn the rich motion prior from a sequence of complete parametric models of posed human body shape. 
Our prior can easily estimate poses in missing frames or noisy measurements despite significant occlusion by employing a temporal attention mechanism. 
More interestingly, our prior can guide the system with incomplete and challenging input measurements to quickly extract critical information to estimate the sequence of poses, significantly improving the training efficiency for mesh sequence recovery. 
ReMP consistently outperforms the baseline method on diverse and practical 3D motion data, including depth point clouds, LiDAR scans, and IMU sensor data.
Project page is available in \href{https://hojunjang17.github.io/ReMP}{https://hojunjang17.github.io/ReMP}.
\end{abstract}

\input{1_introduction}

\input{2_relatedWorks}

\input{3_method}

\input{4_experiments}

\input{5_discussion}

\input{6_conclusion}

\paragraph{Acknowledgments}
This work was supported by the NRF grant (No. RS-2023-00218601) and IITP grant  [No. RS-2021-II211343, Artificial Intelligence Graduate School Program (Seoul National University)] funded by the Korea government(MSIT), and Creative-Pioneering Researchers Program through Seoul National University, Young Min Kim is the corresponding author.

{\small
\bibliographystyle{ieee_fullname}
\bibliography{egbib}
}

\end{document}


\title{Supplementary Material\vspace{0.5em}\\ReMP: Reusable Motion Prior for Multi-domain\\3D Human Pose Estimation and Motion Inbetweening}

\author{Hojun Jang$^{1}$ and Young Min Kim$^{1,2}$ \\
\\
$^1$ Dept. of Electrical and Computer Engineering, Seoul National University\\
$^2$ Interdisciplinary Program in Artificial Intelligence and INMC, Seoul National University\\
{\tt\small \{j12040208, youngmin.kim\}@snu.ac.kr}
}
\maketitle
\section{Implementation Details}

\subsection{Motion Parameters}

We define our motion parameter to be a set of 6D pose parameter and the expanded root translation transition as written below:
\begin{equation}
    M=\left[\theta_{6D}^{\text{flat}}, \underset{3\rightarrow144}{\text{MLP}}(\Delta x)\right]\in \mathbb{R}^{144+144}
\end{equation}
Using 6D pose parameters is widely known to help the network to train and inference the rotation values~\cite{6d_rot}, while using a root transition instead of the absolute root position value and even expanding its dimension is not a common sense.
To justify our choice, we conducted an experiment and compare motion autoencoding performance of the motion prior when the root translation type ($x$ or $\Delta x$) and the dimension (3 or 144) change.
Table~\ref{table:supp_translation} verifies our choice of using the dimension-expanded $\Delta x$ by showing that the reconstruction errors are the best among all choices.
Increasing the dimension of $\Delta x$ enhanced the motion autoencoding performance especially for the root translation estimation.

\subsection{Synthetic Dataset Generation}
We report additional implementation details of the synthetic dataset generation process.

\paragraph{Point Clouds}
To create a synthetic point cloud dataset, we use the Open3D~\cite{open3d} library, which provides various functions for handling 3D data.
First, we generate synthetic depth images of a moving SMPL~\cite{smpl} mesh.
We position a virtual camera by randomly rotating it around a vertical axis and orient it to face the root position at the midpoint of the motion sequence.
After setting the camera’s position and orientation, we capture depth images of the SMPL mesh sequence at a resolution of 640 × 480. 
We then generate a depth point cloud consisting of 1,024 points from these depth images.

For the synthetic LiDAR scan dataset, we utilize the same depth images used for generating depth point clouds.
To simulate the sparsity characteristic of LiDAR scans, we downsample the depth images by taking every fifth row and column.
From these sparse depth images, we randomly sample points to create LiDAR scans, each consisting of 256 points.

\input{tables/supp_translation}

\paragraph{IMUs}
For the synthetic IMU sensor dataset, we follow the procedure outlined in DIP~\cite{DIP-IMU}.
We begin by attaching six virtual IMU sensors to specific vertices of the SMPL~\cite{smpl} mesh.
The attachment points are the left arm (vertex: 1962), right arm (vertex: 5431), left leg (vertex: 1096), right leg (vertex: 4583), head (vertex: 412), and root (vertex: 3021).
Next, we animate the motion sequence to record the acceleration and orientation of the virtual sensors.
The generated sensor data are then paired with the SMPL parameters to form a synthetic dataset.


\subsection{Hyperparameter Setup}

We show the hyperparameter settings we used to train ReMP.
Tables \ref{table:supp_hparams_prior} and \ref{table:supp_hparams_reuse} show the hyperparameter setup for the motion prior training and the reusing prior, respectively.

$N_\text{epoch}$ is the total epoch of the training and when the training reaches the epoch of $N_\text{decay,1}$ or $N_\text{decay,2}$, the learning rate $lr$ decreases to $lr/4$ and $lr/10$, respectively.
We use a sequence of time length $T$ to be 40 and mask out the sequence with the ratio of $r_\text{mask}$ using a \texttt{key\_padding\_mask} in the transformer~\cite{transformers} encoder.
$D_{\text{Tr},z}$, $D_{\text{Tr},\text{ff}}$, $N_{\text{Tr},\text{layer}}$, and $N_{\text{Tr},\text{head}}$ refer to the intermediate latent dimension, feedforward size, number of layers, and number of heads in the transformer network, respectively.
The rest follow the notations in the main paper.

\input{tables/supp_hparams_prior}
\input{tables/supp_hparams_reuse}

\section{Experiment Details}

\subsection{Dataset}
We use AMASS~\cite{amass} to train our model.
AMASS dataset is a large-scale dataset which contains SMPL parameters of more than 20 different datasets, including CMU dataset~\cite{CMU}.
Since we test ReMP on the synthetic CMU dataset, we use the rest to train the motion prior and also the reusing part.
Our motion prior learns the motion within 40 frames at 10 fps, so we split the sequence which is longer than 8 seconds into several pieces and drop the sequence shorter than 4 seconds to make each sequence to be minimum 4 seconds long.
Therefore, the number of sequence we used to train our model is 17,240 and the number of sequence in the synthetic CMU dataset is 2,962.

\subsection{Baselines}
We use the same baselines in point cloud input scenarios for both depth images or LiDAR scans.
VoteHMR~\cite{votehmr} addresses challenges from occlusion and measurement noise in single-view point cloud measurements by segmenting the input point cloud into parts classified as different joints.
Additionally, it requires point segmentation for training, introducing dependencies that our method does not require.
Zuo \etal~\cite{unsup_3d} reconstruct human body mesh surfaces from point cloud inputs.
It first regresses parameters with a neural network and then refine them through optimization employing probabilistic self-supervised loss functions.
The optimization step enhances robustness to outliers but incurs significant computational overhead.
Both methods estimate pose parameters for individual frames, lacking temporal coherence.
In contrast, Jang \etal~\cite{DMR} concurrently regress parameters for a temporal sequence, leveraging temporal information for more accurate and smooth motion.
However, it does not employ efficient encoding schemes and cannot be applied to different sensor modalities.

DIP-IMU~\cite{DIP-IMU} is the first deep learning-based method for human pose estimation from IMU inputs, using a bidirectional RNN architecture to estimate pose parameters, but lacks the ability to estimate root translation.
TransPose~\cite{TransPose} and PIP~\cite{PIP} enhance results by incorporating physical constraints to recover human motion, which enables root translation estimation.
TransPose estimates foot-ground contact probability and root joint velocity, while PIP includes a physics-aware motion optimizer that refines motion using a torque-controlled floating-base simulated character model and a proportional-derivative (PD) controller for better tracking accuracy and physical plausibility.
Every IMU baseline excludes shape parameter estimation, as the IMU sensor input does not contain an information about body shape.
Therefore, we also use ground truth shape parameter for the experiments on IMU sensors.

\section{Additional Results}
We provide additional results of every experiment we conducted with the supplementary video.
Since we focus on the motion, videos show the result more effectively, offering better views to compare with the baselines.

{\small
\bibliographystyle{ieee_fullname}
\bibliography{egbib}
}

%% file: 1_introduction.tex
\section{Introduction}

\input{figures/teaser}

Foundation models have recently demonstrated the power of large-scale datasets, demonstrating innovative results in diverse downstream tasks. However, compared to natural language processing~\cite{bert, gpt3, gpt3.5} (trained with 350 GB-45 TB of data), or image processing~\cite{clip,dino,sam} (trained with 10-400 million images), the counterpart of human motion has remained less explored.
Although we may not be able to collect an internet scale of data with accurate human motion, we argue that existing motion data, such as AMASS~\cite{amass} ($\sim$ 11,000 motions), are sufficient to learn a strong motion prior.
Compared to the possible space of image pixels or arrangement of words, the space of possible human body poses is relatively more constrained and can be successfully modeled with a handful set of parameters~\cite{smpl}.
If we consider the temporal sequence of motion, only a subset of pose sequences is physically plausible and realistic, confined by the human body's skeletal structure and natural capabilities.

Our prior deliberately focuses on accurately reconstructing human mesh sequences with 3D dynamic information.
The prior knowledge can be adapted to other modalities and various downstream tasks that require the capture of 3D motion.
In contrast to the abundant literature that estimates poses in 2D images or video inputs~\cite{2dhuman_1, 2dhuman_2, 2dhuman_3, 2dhuman_4, 2dhuman_5}, our prior is composed of complete 3D configurations that are free from projective distortion, occlusion, or appearance changes.
We also examine a temporal sequence of motion instead of estimating frame-wise poses, which should incorporate temporal context, increasing the robustness of estimation in various adversaries. 
The 3D action space with temporal context can better represent human intents and gestures or temporally predict subsequent motions, enabling us to develop applications that accurately capture human motion or provide necessary services.

In this paper, we obtain a Reusable Motion Prior (ReMP), which contains a tight correlation in 3D temporal sequences from existing datasets~\cite{amass}.
We take a sequence of motion and use a temporal transformer followed by a variational autoencoder (VAE) architecture to obtain a latent distribution. 
The comprehensive parameters provide the geometric variations within the movement, while the transformer applies attention to temporal dependencies. 
Our training formulation is specifically designed to capture fine-grained motion variations within the sequence of motion. 
Specifically, we adapt continuous latent representation to maximize expressivity and maintain a temporal sequence of latent embeddings that correspond to individual time stamps for the encoded sequence. 
We apply random masking in the transformer during training to promote modeling temporal context.
We also incorporate effective input encodings, such as 6D rotation representation, incremental translation, and velocities, and allow the neural network to comprehend detailed dynamics.
Compared to generative priors, our latent space can accurately track the full 3D mesh of human poses.

Once we obtain a comprehensive prior from complete 3D data, we can reuse the prior to extract essential information from various measurement characteristics.
We propose a distillation method that adapts the prior to estimate plausible 3D motion even under noisy, incomplete data or fill the in-between motions given sparse poses.
Even when the data is occluded or sparse, our motion prior encourages the network to detect critical features and quickly train the network to perform motion estimation.
Our work can also regress motion parameters in other modalities that measure dynamic human movements, such as IMU, and show robust performance to estimate accurate movement. 
The aid of pretrained motion prior additionally enhances efficiency and results in better performance than the baselines in smaller training datasets.

In summary, our contributions are threefold:
\begin{itemize}
    \item We introduce ReMP, a novel framework for learning reusable motion prior from large-scale 3D motion dataset, which significantly enhances accuracy and robustness in human pose estimation and motion inbetweening tasks.
    \item We demonstrate an effective formulation for 3D motion prior that can accurately encode dynamic dependencies inherent in human motion sequences that can increase the data usage efficiency. 
    \item We present remapping techniques of our pretrained motion prior to different measurements or various adversaries, demonstrating its versatility and efficiency across diverse motion capture modalities.
\end{itemize}

%% file: figures/teaser.tex
\begin{figure}[t]
\begin{center}
\includegraphics[width=\linewidth]{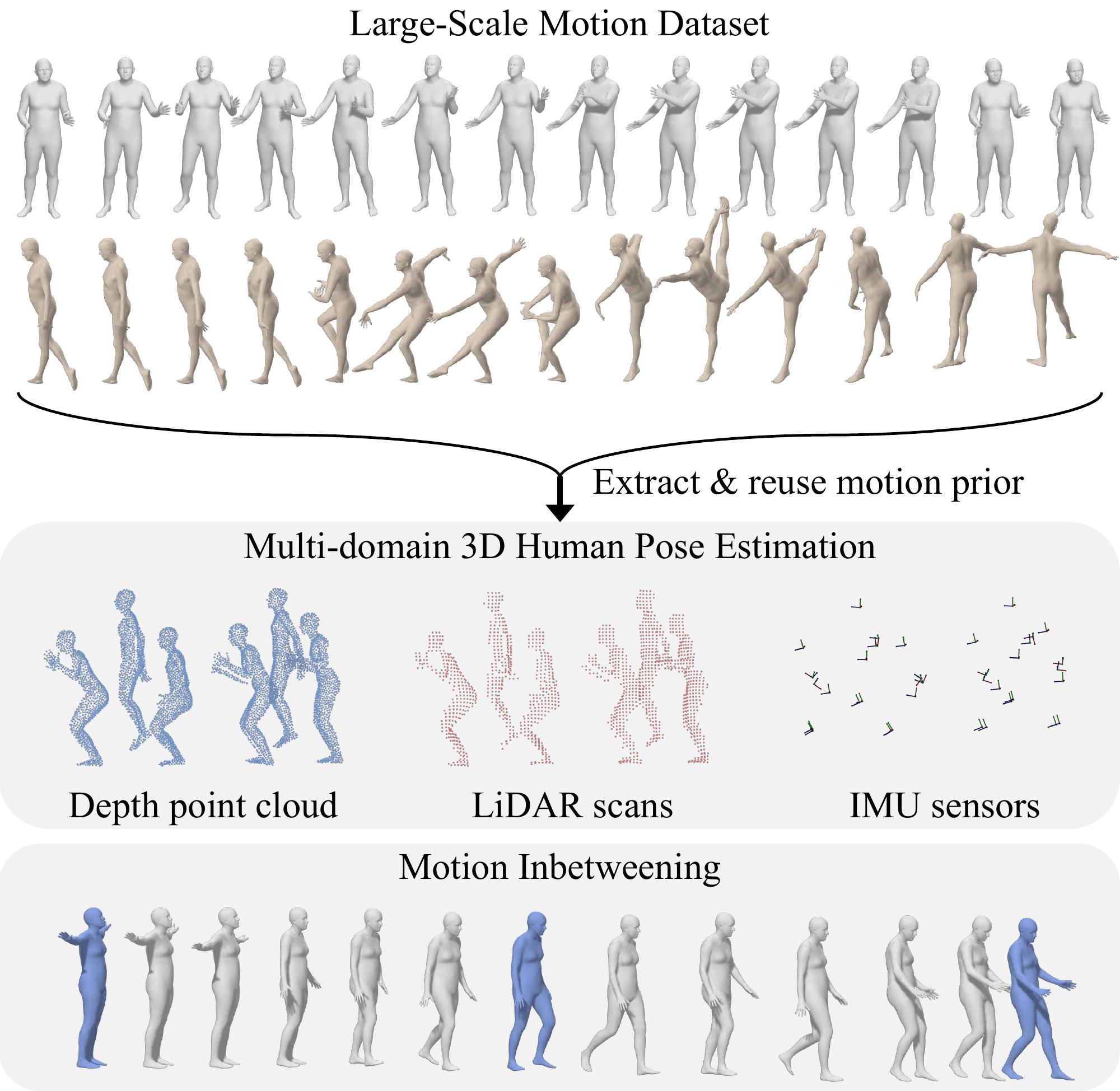}
\vspace{-1em}
\caption{
    We extract rich motion priors from the large-scale motion dataset and reuse them for various applications, such as 3D human pose estimation and motion inbetweening.
}
\label{fig:teaser}
\end{center}
\vspace{-2em}
\end{figure}

%% file: 2_relatedWorks.tex
\section{Related Works}
\subsection{Motion Prior}
Similar to the success of foundation models, many works in human motion understanding leverage large-scale datasets to enhance performance in various tasks. 
While there exists rich literature that finds a latent space for individual poses, we specifically focus on works that incorporate temporal correlations between adjacent frames within a motion sequence.
Motion VAE~\cite{motionvae} and HuMoR~\cite{humor} are notable examples, which employ a conditional variational autoencoder (CVAE) architecture to learn generative priors of the future frame conditioned on the current frame. 
MotionVAE leverages the prior to generate goal-directed human motion within a reinforcement-learning framework, while HuMoR initializes poses from the prior and further refines them through heavy optimization steps.

Other works capture rich dynamic information from a temporal span of motion sequences longer than two consecutive frames.
To avoid posterior collapse that is prevalent with generative priors, additional information often provides structural guidance  in the latent space. 
Actor~\cite{actor} learns a prior combined with action labels, such that a latent code can be generated conditioned on action labels or text descriptions.
PoseGPT~\cite{posegpt} and MoMask~\cite{momask} map motion sequences to a discrete latent space using a vector-quantized variational autoencoder (VQVAE)~\cite{vqvae} architecture, which can produce a sequence of motions by estimating an index of the codebook.

In contrast, our proposed method learns more expressive continuous latent space and mainly focuses on correctly reconstructing 3D full-body motion.
Our priors are exclusively trained from motion sequences without relying on action labels of text descriptions. 
Jang \etal~\cite{DMR} demonstrated that spatio-temporal prior can effectively recover full-body mesh sequences from point cloud input.
Our work learns a more general prior extracted from a parametric model. The motion prior from ReMP is more comprehensive and flexible, allowing it to understand 3D human motion from diverse input modalities.

\subsection{3D Human Pose Estimation}
\label{subsec:3d_human_pose_estimation}

Estimating an accurate human motion from diverse sensor data has gained significant attention due to a wide range of possible applications, such as lightweight motion capture, human-machine interaction, and human behavior analysis. 
Our proposed prior includes the temporal context of 3D motion and can enhance the accuracy of pose estimations in various contexts with diverse sensor modalities.
Previous works rely on tracking devices with carefully calibrated cameras, or individual frames are processed separately.
Even when using more casual input modalities, most previous works require separate training of the model for different tasks or input modalities.

\paragraph{Depth Camera}
A depth camera is one of the most practical inputs to estimate 3D human motion.
While the data may suffer from occlusions and measurement noises, commodity devices can quickly obtain 3D measurements of humans without further calibration or markers.
A large volume of previous works processes individual frames to obtain a reasonable estimate, which leads to unnatural jittering in estimated poses~\cite{3d-coded, votehmr, unsup_3d}.
Jang~\etal~\cite{DMR} increases the time span to longer motion sequences and directly estimates the motions without optimization in point cloud input.
ReMP further demonstrates incorporating a coherent motion prior to different input modalities.

\paragraph{LiDAR}
LiDAR is one of the most widely used 3D sensors, especially for autonomous driving, and there is a demand for accurate human motion recognition in the data.
While depth cameras and LiDAR both capture point cloud measurements, LiDAR measurements are significantly sparser than depth camera outputs, making human pose estimation from LiDAR data more challenging.
Furthermore, few LiDAR datasets are available to train reliable human pose estimation models.
 Some LiDAR scan datasets contain human motion~\cite{waymo, nuscenes} but lack accurate ground truth pose annotations.
Some works provide LiDAR scan to SMPL annotations~\cite{lidarcap, humanm3, sloper4d}, but they contain limited data and may lack the quality needed to train accurate pose estimation models.
By leveraging reusable motion prior and our synthetically generated dataset, we can estimate accurate temporal motion in challenging LiDAR datasets, overcoming the limitations of sparse data and the lack of comprehensive ground truth annotations.

\paragraph{IMU Sensor}
IMU sensors, commonly used in AR/VR and wearable devices, are cost-effective devices that can be directly attached to body parts to reconstruct human motion.
SIP~\cite{sip} was among the first to utilize IMU sensor data for human pose estimation, optimizing pose parameters to match the input sensor data.
Subsequent works, such as DIP-IMU~\cite{DIP-IMU}, employed deep learning architectures like RNNs for pose estimation.
Recent advancements, including \cite{TransPose, PIP, tip}, have enhanced performance by integrating physics-based constraints.
ReMP leverages a pretrained motion prior from a large-scale motion dataset, which leads to physically plausible motion sequences that adhere to the physical constraints of the human body.

%% file: 3_method.tex
\section{Method}

\input{figures/pipeline}

We encode the full 3D mesh of diverse human poses with SMPL~\cite{smpl}.
SMPL is the most widely used parametric human model to map the full-body mesh composed of 6890 vertices $V\in\mathbb{R}^{6890\times3}$ into a set of parameters.
The parameters include pose parameters $\theta\in \mathbb{R}^{24\times3}$, root translation $x\in\mathbb{R}^{3}$, and shape parameters $\beta\in\mathbb{R}^{10}$.
We convert the pose parameter from an axis-angle representation to a 6D rotation representation $\theta_{6D}\in\mathbb{R}^{24\times6}$, which is more suitable for training neural networks and resolves the ambiguity issue~\cite{6d_rot}.
Following \cite{humor}, we use the difference in root translation between consecutive frames ($\Delta x$) in the global coordinate system rather than the absolute value.
Na\"ive concatenation of translation to pose parameters often overlooks discrepancies due to dimensional imbalance.
To address this, we expand the dimensionality of $\Delta x$ using a simple MLP to match the dimensions of $\theta_{6D}$, and then concatenate them to form the motion parameters for a single pose
\begin{equation}
    M=\left[\theta_{6D}^{\text{flat}}, \underset{3\rightarrow144}{\text{MLP}}(\Delta x)\right]\in \mathbb{R}^{144+144}.
\end{equation}
Then, we represent human motion using a sequence of motion parameters $M_{1:T}$.
We validate the use of $\Delta x$ instead of the absolute translation value and the dimension expansion to $\Delta x$ in the supplementary material.

Our method consists of two main steps as described in Figure~\ref{fig:pipeline}.
In the first phase, we train motion prior and learn the temporal context of the motion parameter sequence (\cref{sec:motion_prior}).
The second reusing phase learns to perform various tasks using the trained motion prior (\cref{sec:reuse_prior}).

\subsection{Training Motion Prior}
\label{sec:motion_prior}
We train the motion prior with a large-scale motion dataset containing diverse motions.
At a high level, our approach follows the regular variational autoencoder (VAE)~\cite{vae} training scheme.
We employ a continuous latent space to express fine-grained motion accurately.
We use a VAE architecture based on a transformer network~\cite{transformers} as the backbone.
The transformer encoder takes the motion parameters $M_{1:T}\in\mathbb{R}^{T\times288}$ as input to create a motion feature $z'_{1:T}\in\mathbb{R}^{T\times D_{z}}$.
Each transformer input channel takes a single motion parameter, with positional embeddings added, to effectively learn the temporal correlations of the motion.
The motion sequence of length $T$ is embedded into a sequence of continuous latent variables of the same length, allowing a sufficiently rich latent space to capture fine-grained motion variations.
Using the motion feature vector, we employ separate 2-layer MLPs to generate a posterior distribution $q_\phi \left( z_t | M_t, z'_t \right)$ and a prior distribution $p_\psi \left( z_t | z'_t \right)$, where $z_t\in\mathbb{R}^{D_z}$ is a latent vector at the $t^{th}$ time step.
We assume Gaussian distributions for both the posterior and prior distributions.
After sampling $z_{1:T}$ from these distributions, we input the latent vectors into the transformer decoder to reconstruct the SMPL parameters.

Our reconstruction loss covers comprehensive aspects of motion reconstruction.
Specifically, we use the following loss terms to train the posterior distribution:
%
\begin{equation}
\label{eq:posterior}
    \mathcal{L}_{\text{recon}}=\sum_{s\in S}{w_s \mathcal{L}_s},\ \  S=\left\{ \theta, \Delta\theta, x, \Delta x, J, V \right\}
\end{equation}
$\mathcal{L}_{\theta}$ is a pose parameter difference loss, which is L2 distance between the estimated and the ground truth 6D pose parameters.
$\mathcal{L}_{\Delta \theta}$ is an angular velocity loss, which makes the angular velocity of the estimated pose parameter match the ground truth.
$\mathcal{L}_{x}$ and $\mathcal{L}_{\Delta x}$ are the loss terms that make the estimated root translation position and its speed to be similar to the ground truth, respectively.
Finally, $\mathcal{L}_{J}$, and $\mathcal{L}_{V}$ are the L2 distance between the estimated joints and vertices to be near the ground truth, respectively.
Additionally, we encourage the prior distribution to follow the posterior distribution using a KL divergence loss:
\begin{equation}
    \mathcal{L}_{\text{KL}}^{\text{prior}}=\frac{1}{T} \sum_{t}{D_{KL}\left( q_\phi \left( z_t | M_t, z'_t \right) \| \ p_\psi \left( z_t | z'_t \right) \right)}.
\end{equation}
The total loss function to train motion prior is as follows:
\begin{equation}
    \mathcal{L}_{\text{prior}}=\mathcal{L}_{\text{recon}} + w_{\text{KL}}^{\text{prior}} \mathcal{L}_{\text{KL}}^{\text{prior}}.
\end{equation}

For better generalizability, we randomly mask some frames of the input motion parameters by utilizing a \texttt{\small key\_padding\_mask} in the transformer encoder.
The random temporal mask helps our model to better encode the unseen motion sequence and increase robustness to our motion autoencoder.
Thanks to the use of input masking, our model can also perform motion inbetweening task.

\subsection{Reusing Pretrained Prior}
\label{sec:reuse_prior}

After training the motion prior with the large-scale dataset, we freeze the networks and reuse the rich motion prior to reconstructing motions for different input conditions.
Our reusing phase mainly consists of two parts: an input encoder that processes different inputs, followed by a latent mapper, which brings the encoded input into the pretrained latent space from the first phase.
These transforms can be quickly trained with relatively a small amount of data, even with synthetic data, and such that we can effectively exploit the motion context of reusable prior, which is further demonstrated in \cref{sec:experiments}.

As the observation from a different sensor modality comes in, the input encoder encodes the data $I_{1:T}$ to a meaningful feature $I'_{1:T}\in\mathbb{R}^{T\times D_I'}$.
We use PointNet~\cite{pointnet} as the input encoder for the point cloud input and 2-layer MLP for the IMU data input.
The encoded features are then put into the latent mapper, which is also a transformer encoder.
The latent mapper generates the Gaussian distribution $p'_\psi \left( z_t | I'_t \right)$ that can be used to sample the latent vectors from the input features.
The distribution is forced to follow the distribution of the prior distribution $p_\psi$ by a KL divergence loss
\begin{equation}
    \mathcal{L}_{\text{KL}}^{\text{reuse}}=\frac{1}{T} \sum_{t}{D_{KL}\left( p_\psi \left( z_t | z'_t \right) \| \ p'_\psi \left( z_t | I'_t \right) \right)}.
\end{equation}
Also, for better estimation performance, we leverage the same loss terms used to train the posterior distribution $q_\phi$ as shown in \cref{eq:posterior}.

As motion prior does not contain information about the shape parameters $\beta$, we have an additional shape parameter estimator.
Shape parameter estimator is a 2-layer MLP that receives an encoded input feature $I'_{1:T}$ as an input and outputs an estimate $\hat{\beta}$.
The shape parameter should not change during the motion, so we flatten $I'_{1:T}$ into a one-dimensional vector and put it to the shape estimator MLP to generate a single $\hat{\beta}$ for each sequence.
To train the shape estimator, we use a weighted loss function that penalizes more to the former part of $\beta\in\mathbb{R}^{10}$ as the shape parameters are the result of PCA.
That is, we use our shape loss to be
\begin{equation}
    \mathcal{L}_{\beta} = \sum_{i=1}^{10}{ \frac{11-i}{10}  \left(\beta_i - \hat{\beta}_i \right) ^ 2}.
\end{equation}
Therefore, the final loss function to train reusing the prior is
\begin{equation}
    \mathcal{L}_{\text{reuse}}=\mathcal{L}_{\text{recon}}+w_{\text{KL}}^{\text{reuse}} \mathcal{L}_{\text{KL}}^{\text{reuse}}+ w_{\beta} \mathcal{L}_{\beta}.
\end{equation}

%% file: figures/pipeline.tex
\begin{figure*}[t]
\begin{center}
\includegraphics[width=0.9\linewidth]{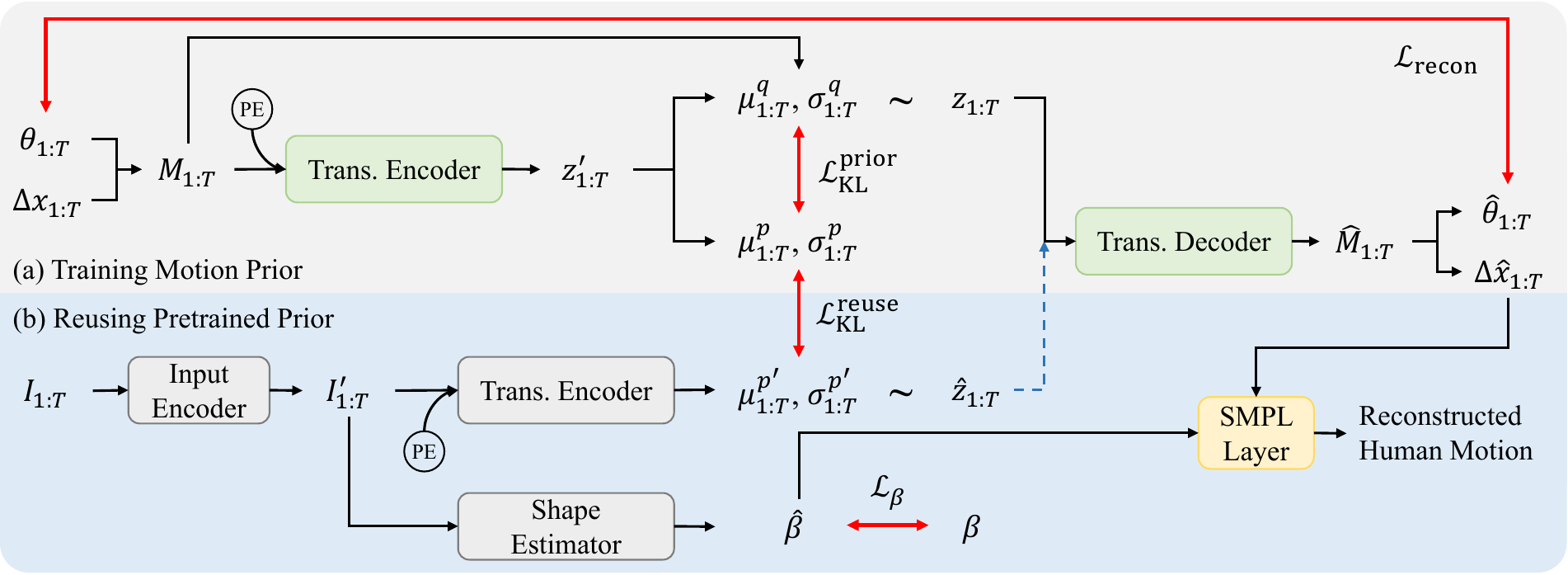}
\caption{
    The overall pipeline of our method consists of two parts: (a) training motion prior and (b) reusing pretrained prior.
    In the motion prior training phase, a sequence of pose parameters $\theta_{1:T}$ and the root translation transitions $\Delta x_{1:T}$ form a sequence of motion parameter $M_{1:T}$.
    We use a transformer encoder and MLP layers to generate Gaussian distributions where we can sample the latent vectors.
    We feed the latent vectors to a transformer decoder to generate the motion parameters then to the SMPL parameters.
    After training the prior, we freeze all the networks used in the first phase.
    In the reusing phase, we encode the input data and use a transformer encoder to generate a distribution that is then used to sample the latent vectors for the transformer decoder.
    We use an additional shape parameter estimator for $\beta$.
    Finally, we combine all three parameters with the SMPL layer to reconstruct the human motion.
}
\label{fig:pipeline}
\end{center}
\vspace{-1em}
\end{figure*}

%% file: 4_experiments.tex
\section{Experiments}
\label{sec:experiments}

This section demonstrates how the prior motion is reused for downstream tasks.
First, we present the results of ReMP, showing its superior performance over baselines in versatile 3D human pose estimation tasks in various sensor modalities (\cref{subsec:motion_est}) and motion inbetweening (\cref{subsec:interpolation}).
We also highlight the effectiveness of motion prior in \cref{subsec:exp_prior}.

\paragraph{Synthetic Dataset Generation}
Our outputs require accurate SMPL parameters for supervision using $\mathcal{L}_\text{recon}$.
However, datasets containing both input sensor data and accurate SMPL parameters are rare.
We, therefore, first generate training datasets for different input modalities, following the method introduced in the previous works~\cite{surreal, DMR, DIP-IMU}.
We utilize the AMASS dataset~\cite{amass} for both motion prior training and the reusing phases of our method.
The AMASS dataset is a comprehensive human motion dataset comprising approximately 11,000 motion sequences with SMPL parameters, providing ample data to learn rich motion priors.

For the reusing phase, we generate synthetic datasets for each input modality.
Using the Open3D library~\cite{open3d}, we render SMPL meshes and generate depth images, which are then converted into point clouds.
We create dense depth images for synthetic depth point clouds and sparse depth images for LiDAR point clouds, using 1,024 and 256 points, respectively.
For the IMU dataset, we attach six synthetic IMUs to the SMPL mesh and track the acceleration and orientation of each sensor during the movements.
Each sequence is generated with 40 frames at a framerate of 10 fps.
We show additional details of the data generation process in the supplementary material.

After synthetic training, we can directly test our method on real-world datasets.
The synthetic datasets provide accurate 3D motion information, which is robust and effective in practical scenarios.

\subsection{Motion Estimation}
\label{subsec:motion_est}

We compare the performance of 3D human pose estimation against state-of-the-art methods for depth point cloud, LiDAR, and IMU inputs.
The supplementary material contains descriptions on the baseline methods and additional results with videos.
ReMP excels in human pose estimation in all scenarios without the need for additional information or heavy optimization steps, demonstrating robust and versatile performance across different tasks.

\paragraph{Depth Point Cloud}
\input{tables/depth_results}
\input{figures/depth_results}

We evaluate the 3D human pose estimation performance of ReMP and the baselines~\cite{unsup_3d, votehmr, DMR} trained on the synthetic AMASS dataset.
Firstly, we test ReMP on the synthetic CMU~\cite{CMU} dataset, which is part of the larger synthetic AMASS dataset we generated.
This dataset was not seen during training.
Secondly, we evaluate ReMP on Berkeley MHAD dataset~\cite{BerkeleyMHAD}, captured using a Microsoft Kinect sensor~\cite{kinect}, to demonstrate the applicability of ReMP on real-world data.
Given that the Berkeley MHAD dataset is an unsegmented depth image including surrounding environments, we preprocess it by roughly cropping the points near the actor using a bounding box.

We use four time-averaged metrics to evaluate the performance.
Pose error is the angular difference between the ground truth and the estimated pose parameters.
Joint and mesh errors are the Euclidean distance errors of the joints and mesh vertices, respectively.
Finally, Chamfer distance~\cite{chamfer_distance} is used to evaluate the Berkeley MHAD dataset since it does not contain ground truth SMPL parameters.

Table \ref{table:depth} presents the quantitative results of our method compared to baselines on both the synthetic and real datasets.
ReMP consistently outperforms all baselines across every metric, demonstrating its robustness and effectiveness.
Berkeley MHAD dataset highlights ReMP's sim-to-real performance, effectively handling real-world noise without further fine-tuning.
While Jang~\etal~\cite{DMR} also use a motion prior trained with point cloud, the general and comprehensive prior of ReMP results in superior performance.

Figure \ref{fig:depth_results} shows the qualitative results of each method.
VoteHMR~\cite{votehmr} and Zuo \etal~\cite{unsup_3d} struggle to reconstruct continuous motion as they find independent poses for individual frames.
Jang \etal, in contrast, successfully estimates coherent motion throughout the sequence.
However, among all methods, ReMP demonstrates the highest quality in estimated human motion.

\paragraph{LiDAR Scans}
\label{subsec:lidar}
\input{tables/lidar_results}

\input{figures/lidar_results}

We test our method for the LiDAR point cloud on SLOPER4D dataset~\cite{sloper4d}, which is a real LiDAR scan dataset.
Table \ref{table:lidar} shows the quantitative result of our method and the baselines on SLOPER4D, and Figure~\ref{fig:lidar_results} visualizes the results with the error maps.
The data is much more sparse and noisy compared to the depth point cloud, resulting in higher errors compared to Table~\ref{table:depth}.
ReMP shows the best mesh reconstruction result among the pose estimation methods, especially for hand positions.
VoteHMR~\cite{votehmr} and Zuo \etal~\cite{unsup_3d} fail to estimate smooth motion, as in the depth point cloud scenario.

\paragraph{IMU Sensor Data}
\label{subsec:imu}

\input{figures/imu_results}
\input{tables/IMU_results}

After the motion estimation module is trained with synthetic IMU datasets, we evaluate the performance on TotalCapture dataset~\cite{TotalCapture}, which contains real IMU measurements.
We report the performance using five evaluation metrics introduced in DIP~\cite{DIP-IMU}, which are widely used.
SIP error is a mean orientation error of the upper arms and legs in the global space.
Angular error is a mean global rotation error of all body joints.
Positional error and mesh error are the mean Euclidean distance errors of all estimated joints and mesh vertices, respectively, with the root joint aligned.
Finally, a jitter is a mean jerk, a time derivative of the acceleration of all body joints.

We report the results of all methods in Table~\ref{table:imu}.
ReMP outperforms baselines, except for the angular error which is larger than PIP~\cite{PIP}.
Jitter is the most significant enhancement over the baselines, showing that the inherent temporal coherence induces plausible motion estimation.
Figure~\ref{fig:imu_results} emphasizes that ReMP has the best performance in recovering human motion from the IMU input.
We omit the visualization of DIP since it cannot estimate root translation.

\subsection{Motion Inbetweening}
\label{subsec:interpolation}
\input{figures/inbetweening_results}

Motion inbetweening aims to interpolate between two key poses to generate intermediate frames.
While linear interpolation (LERP) and spherical linear interpolation (SLERP) provide straightforward approaches, they may not always yield realistic results, particularly over longer time intervals or periodic motions such as locomotion.
Leveraging motion prior, one can generate multiple plausible motion sequences starting from the initial keyframe and select the one with the closest pose of the ending keyframe pose, as proposed in Neural Marionette~\cite{neural_marionette}.
However, generating and comparing multiple candidates can be computationally inefficient and not guarantee precise in-betweening.
ReMP, on the other hand, utilizes temporal masks while training the transformer encoder, which means we can appropriately adjust the \texttt{\small key\_padding\_mask} to learn motion in-betweening.

Figure \ref{fig:inbetweening_results} shows the result of the motion in-betweening task of ReMP and SLERP, which shows superior results compared to LERP.
The error map indicates that ReMP closely recovers the original motion, while simple interpolation of SLERP fails to reconstruct detailed motion variations.
We do not show the result of Neural Marionette because they use different skeletal structures and only generate output voxels instead of full parametric mesh.

%% file: tables/depth_results.tex
\begin{table}[t]
    \begin{center}
    \resizebox{0.92\linewidth}{!}{
    \begin{tabular}{l|C{1.0cm}C{1.0cm}C{1.0cm}|c}
        \toprule
        \multirow{2}{*}[-0.8em]{Method} & \multicolumn{3}{c|}{Synthetic CMU} & B-MHAD \\
        \cmidrule{2-5}
        & Pose [\unit{\degree}] & Joint [\unit{\centi\meter}] & Mesh [\unit{\centi\meter}]& CD [\unit{\centi\meter}]  \\
        \midrule
        VoteHMR~\cite{votehmr} & 6.71 & 13.76    &   15.20&   28.31  \\
        Zuo \etal~\cite{unsup_3d}    &  7.58   &   14.78  &   16.23 &   31.56  \\
        Jang \etal~\cite{DMR} & 5.43 & 11.15 & 12.62 & 18.71 \\
        \midrule
        ReMP$^\dagger$ & 6.37 & 18.58 & 20.47 & 25.59 \\
        \textbf{ReMP} & \textbf{4.90} & \textbf{9.89} & \textbf{11.16} & \textbf{17.61} \\
        
        \bottomrule
    \end{tabular}
    }
    \end{center}
    \vspace{-1em}
    \caption{
        Comparison with the previous methods, which reconstruct SMPL parameters from depth point clouds, on synthetic (CMU~\cite{CMU}) and real-world (B-MHAD, Berkeley MHAD~\cite{BerkeleyMHAD}) datasets.
        ReMP$^\dagger$ refers to the model directly trained on synthetic AMASS~\cite{amass} without using the motion prior.
    }
    \label{table:depth}
    \vspace{-1em}
\end{table}

%% file: figures/depth_results.tex
\begin{figure*}[t]
\begin{center}
\includegraphics[width=\linewidth]{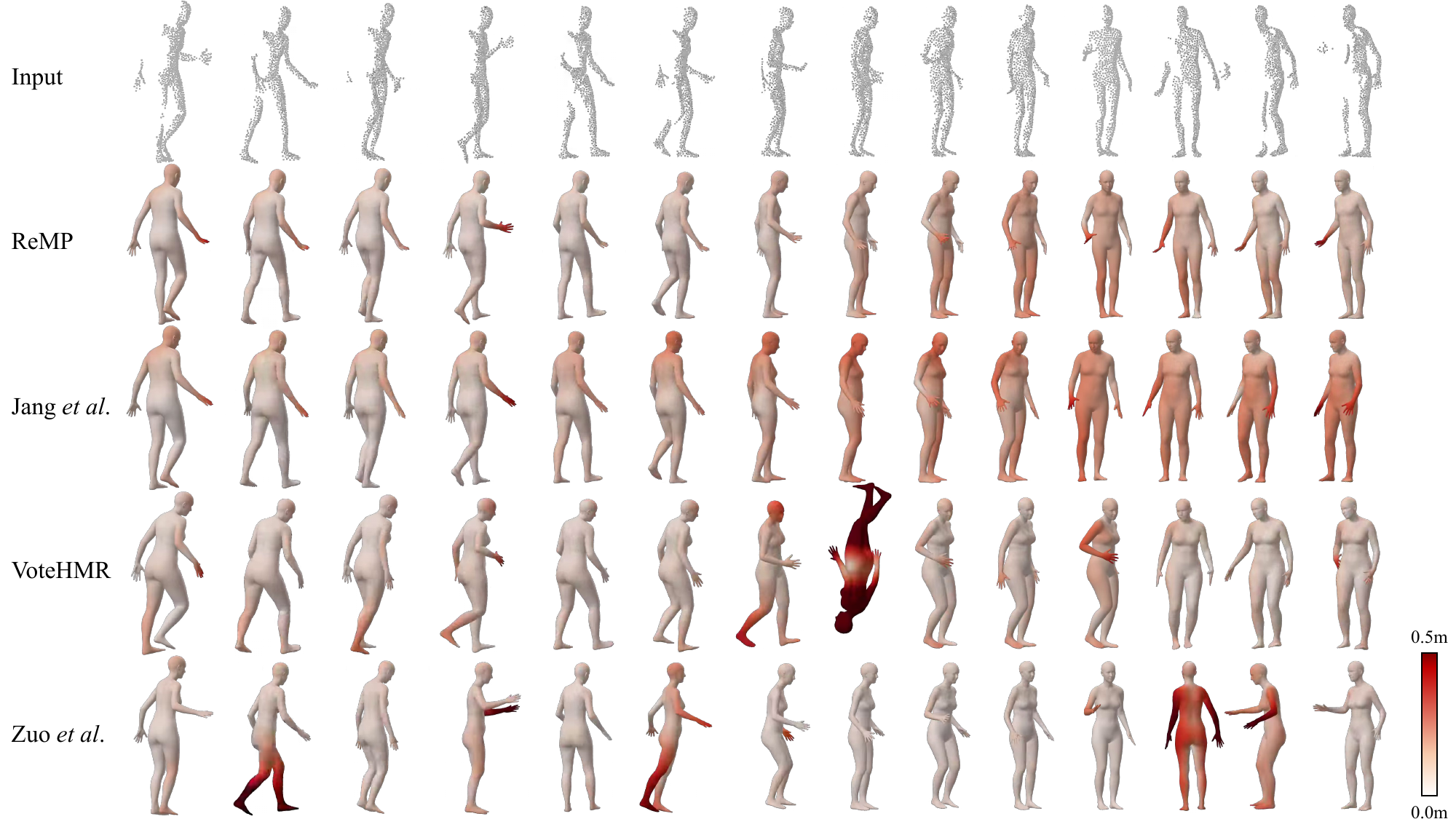}
\vspace{-1.5em}
\caption{
    Results of ReMP and the baselines on synthetic CMU~\cite{CMU} depth point cloud data.
    The colors on the mesh indicate the displacement from the ground truth vertices.
}
\label{fig:depth_results}
\end{center}
\vspace{-1.5em}
\end{figure*}

%% file: tables/lidar_results.tex
\begin{table}[t]
    \begin{center}
    \resizebox{0.95\linewidth}{!}{
    \begin{tabular}{l|C{1.8cm}C{1.8cm}C{1.8cm}}
        \toprule
        \multirow{2}{*}[-0.2em]{Method} & \multicolumn{3}{c}{SLOPER4D}\\
        \cmidrule{2-4}
         & Pose [\unit{\degree}] & Joint [\unit{\centi\meter}] & Mesh [\unit{\centi\meter}] \\
        \midrule
        VoteHMR~\cite{votehmr} & 8.59 & 36.33 & 53.79\\
        Zuo \etal~\cite{unsup_3d} & 10.48 & 43.85 & 43.87 \\
        Jang \etal~\cite{DMR} & \textbf{8.57} & 22.28 & 22.54 \\
        \midrule
        ReMP$^\dagger$ & 9.96 & 87.46 & 96.94\\
        \textbf{ReMP} & 8.58 & \textbf{21.66} & \textbf{22.03} \\
        \bottomrule
    \end{tabular}
    }
    \end{center}
    \vspace{-1.5em}
    \caption{
        Comparison with the previous methods, which reconstruct SMPL parameters from LiDAR scan, on SLOPER4D~\cite{sloper4d} dataset.
        ReMP$^\dagger$ refers to the model directly trained on synthetic AMASS~\cite{amass} without using the motion prior.
    }
    \label{table:lidar}
    \vspace{-1em}
\end{table}

%% file: figures/lidar_results.tex
\begin{figure*}[t]
\begin{center}
\includegraphics[width=\linewidth]{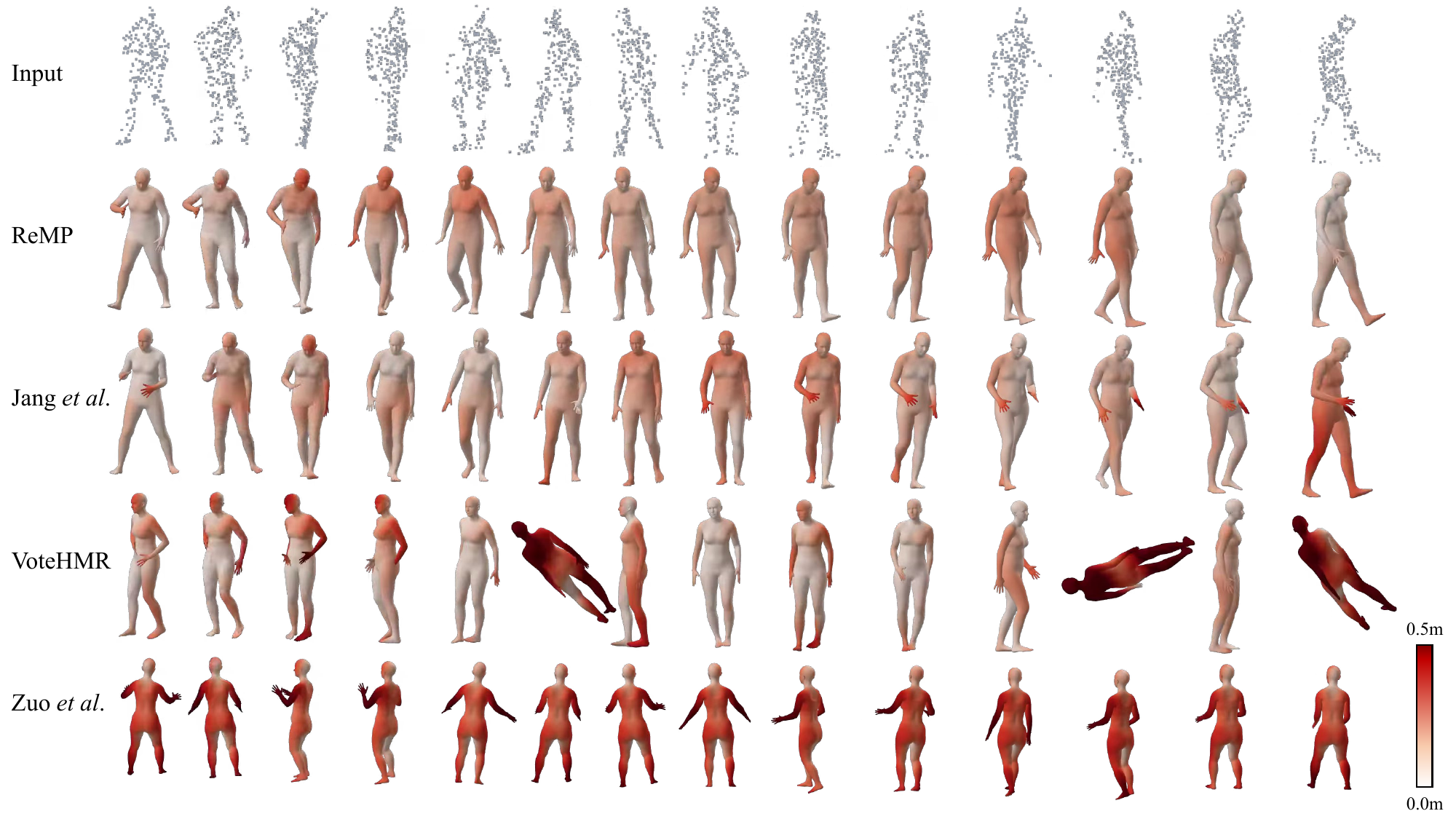}
\vspace{-2em}
\caption{
    Pose estimation results of ReMP and the baselines on SLOPER4D dataset~\cite{sloper4d}.
    The colors on the mesh indicate the displacement from the ground truth vertices.
}
\label{fig:lidar_results}
\end{center}
\vspace{-1.5em}
\end{figure*}

%% file: figures/imu_results.tex
\begin{figure*}[t]
\begin{center}
\includegraphics[width=\linewidth]{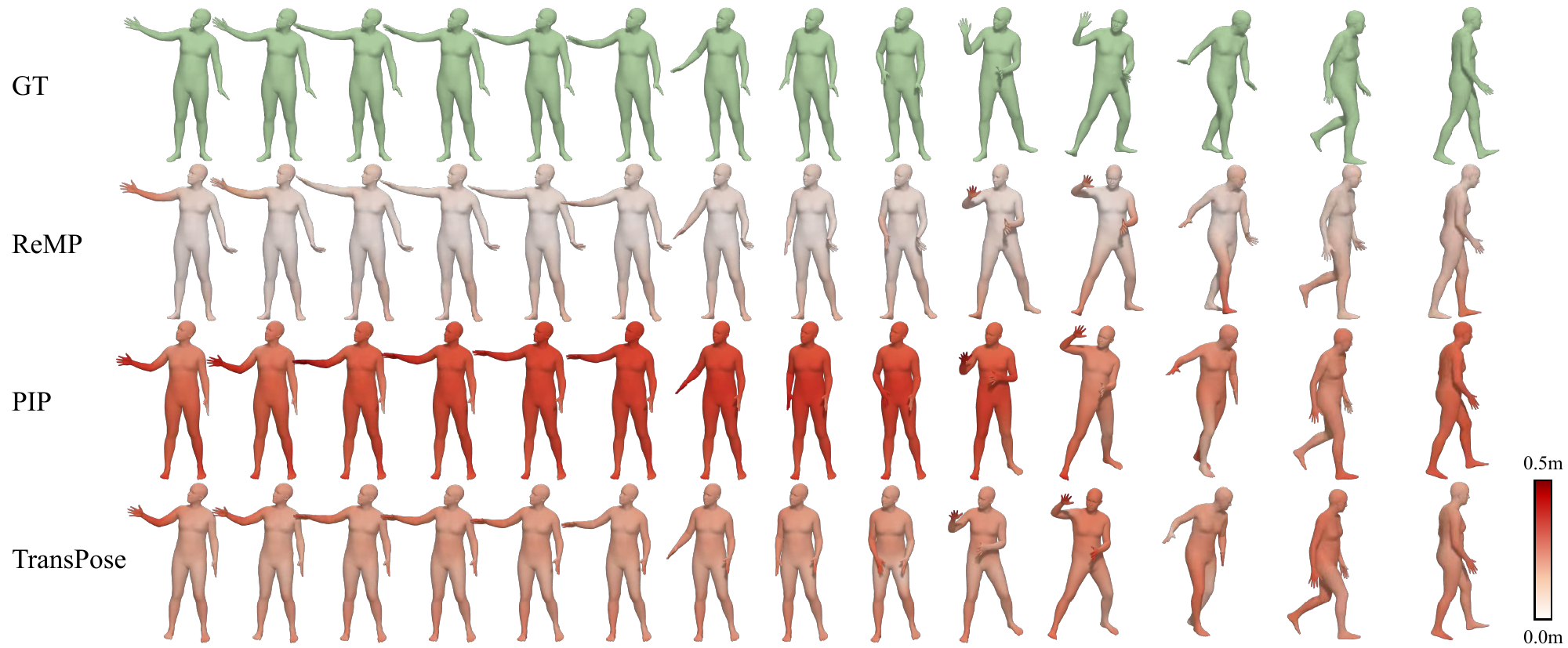}
\vspace{-1.5em}
\caption{
    Motion reconstruction results of ReMP and the baselines from IMU sensor data on TotalCapture dataset~\cite{TotalCapture}.
    The colors on the mesh indicate the displacement from the ground truth vertices.
}
\label{fig:imu_results}
\end{center}
\vspace{-2em}
\end{figure*}

%% file: tables/IMU_results.tex
\begin{table}[t]
    \begin{center}
    \resizebox{\linewidth}{!}{
    \begin{tabular}{l|C{1.0cm}C{1.0cm}C{1.0cm}C{1.0cm}C{1.0cm}}
        \toprule
        \multirow{2}{*}[-0.8em]{Method} & \multicolumn{5}{c}{TotalCapture}\\
        \cmidrule{2-6}
         & SIP \ [\unit{\degree}] & Ang [\unit{\degree}] & Pos [\unit{\centi\meter}] & Mesh [\unit{\centi\meter}] & Jitter [\unit[per-mode=symbol]{km/s^3}] \\
        \midrule
        DIP~\cite{DIP-IMU} & 18.93 & 17.50 & 9.57 & 11.40 & 35.94 \\
        TransPose~\cite{TransPose} & 16.69 & 12.93 & 6.61 & 7.49 & 9.44 \\
        PIP~\cite{PIP} & 12.93 & \textbf{12.04} & 5.61 & 6.51 & 0.20 \\
        \midrule
        ReMP$^\dagger$ & 15.11 & 13.35 & 6.43 & 7.51 & 0.08 \\
        \textbf{ReMP} & \textbf{12.43} & 12.07 & \textbf{5.49} & \textbf{6.35} & \textbf{0.03} \\
        \bottomrule
    \end{tabular}
    }
    \end{center}
    \vspace{-1.5em}
    \caption{
        Comparison with the previous methods, which reconstruct SMPL parameters from IMU sensor data, on TotalCapture~\cite{TotalCapture} dataset.
        ReMP$^\dagger$ refers to the model directly trained on synthetic AMASS~\cite{amass} without using the motion prior.
    }
    \label{table:imu}
    \vspace{-0.5em}
\end{table}

%% file: figures/inbetweening_results.tex
\begin{figure}[t]
\begin{center}
\includegraphics[width=\linewidth]{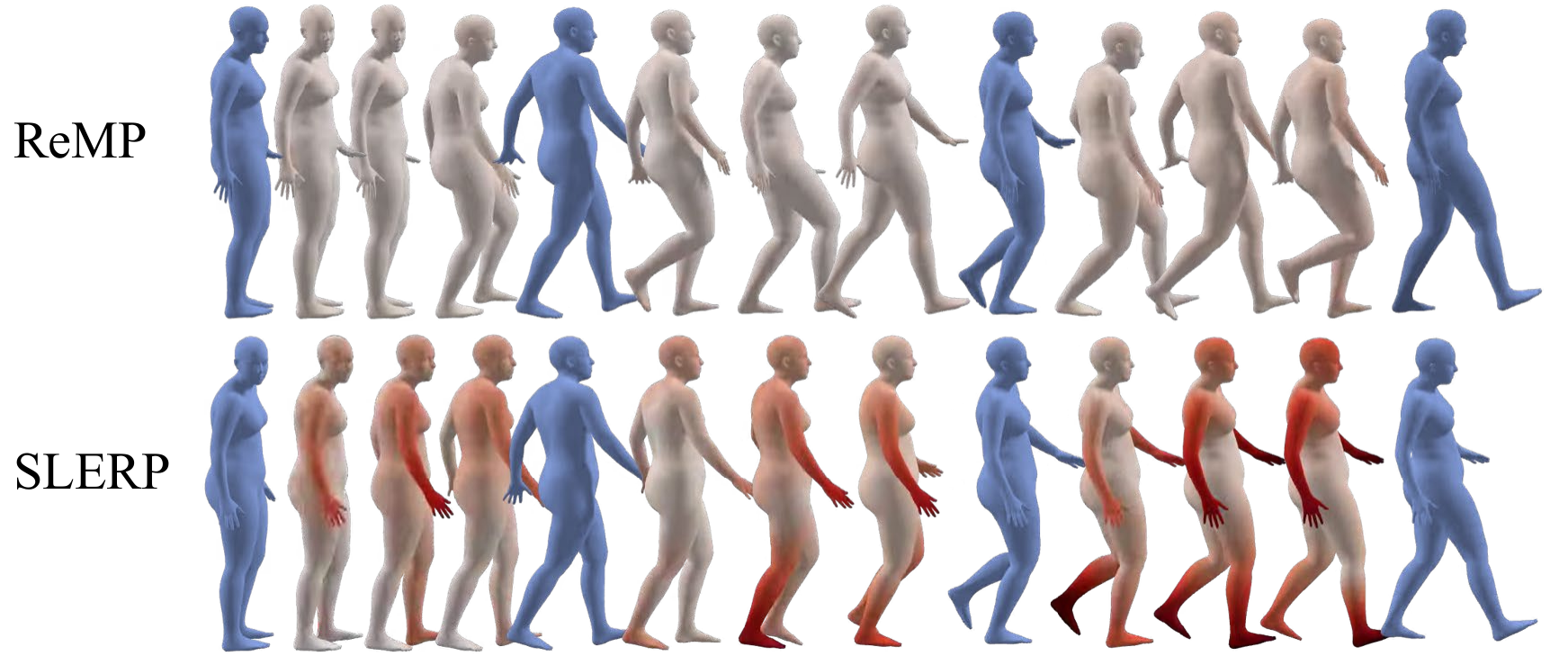}
\vspace{-1.5em}
\caption{
    Motion inbetweening results of ReMP and SLERP when the blue frames are given.
    ReMP better generates the inbetween frames of the sequence while SLERP fails to interpolate realistic motion.
}
\label{fig:inbetweening_results}
\end{center}
\vspace{-2em}
\end{figure}

%% file: 5_discussion.tex
\subsection{Effects of Motion Prior}
\label{subsec:exp_prior}

The foundational motion prior of ReMP guides various tasks of reconstructing fine-grained motions effectively.
To assess the impact of motion priors, we train a variant of our model from scratch in a supervised fashion, without leveraging motion priors, referred to as ReMP$^{\dagger}$.
We report the results in Tables~\ref{table:depth}, \ref{table:lidar}, and  \ref{table:imu}.
In all scenarios, reusing the motion prior significantly boosts the performance of human motion estimation.

\input{figures/data_efficiency}

To further demonstrate the effectiveness of ReMP, we conducted an experiment to evaluate its data efficiency compared to the baselines on synthetic CMU~\cite{CMU} dataset.
We progressively reduce the training data size to 1/2, 1/4, 1/8, and 1/16 of the original dataset and observe resulting pose errors in Figure~\ref{fig:data_efficiency}.
ReMP achieves comparable results even when the training data is reduced to 1/4 of its original size.
Moreover, ReMP maintains superior performance relative to the baselines as the dataset size decreases, underscoring its efficiency in leveraging limited training data.

%% file: figures/data_efficiency.tex
\begin{figure}[t]
\begin{center}
\includegraphics[width=\linewidth]{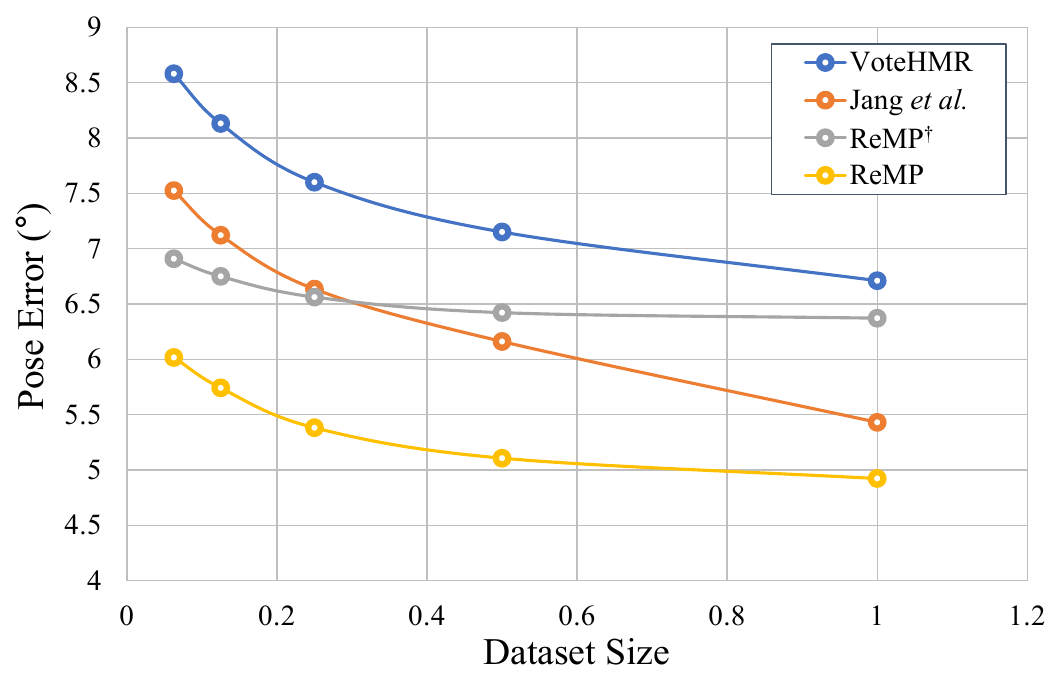}
\vspace{-2em}
\caption{
    Error curve of ReMP and the baselines on differing dataset size.
}
\label{fig:data_efficiency}
\end{center}
\vspace{-1.5em}
\end{figure}

%% file: 6_conclusion.tex
\section{Conclusion}
We presented a reusable motion prior that captures rich temporal correlations of detailed 3D human motion from large-scale datasets.
Leveraging this pretrained motion prior, we achieved superior performance in human pose estimation from various sensor data compared to existing methods.
The module with the reusable motion prior can maintain robust performance in real sensor noises without further fine-tuning and quickly estimate accurate motions with relatively small amount of supervision, even only with synthetically generated data.
Additionally, the temporal information accurately filled in the missing frames of motion sequences.
We expect the versatility of ReMP can be adapted in various  real-world applications requiring fast and accurate 3D human pose estimation under unknown adversaries or unstable network connections.

%% file: tables/supp_translation.tex
\begin{table}[t]
    \begin{center}
    \resizebox{\linewidth}{!}{
    \begin{tabular}{ll|C{1.8cm}C{1.8cm}C{1.8cm}}
        \toprule
        Type & Dim. & Pose [\unit{\degree}] & Trans. [\unit{\centi\meter}] & Joint [\unit{\centi\meter}] \\
        \midrule
        $x$ & 3 & 0.92 & 6.06 & 6.96 \\
        $x$ & 144 & 0.93 & 7.58 & 8.31 \\
        $\Delta x$ & 3 & \textbf{0.91} & 2.92 & 3.88 \\
        $\Delta x$ & 144 & \textbf{0.91} & \textbf{2.26} & \textbf{3.68} \\
        \bottomrule
    \end{tabular}
    }
    \end{center}
    \vspace{-1em}
    \caption{
        Verification of using an expanded $\Delta x$ for our motion parameter.
    }
    \label{table:supp_translation}
\end{table}

%% file: tables/supp_hparams_prior.tex
\begin{table}[t]
    \begin{center}
    \resizebox{0.9\linewidth}{!}{
    \begin{tabular}{lr|lr|lr}
        \toprule
        \multicolumn{2}{c|}{Train} & \multicolumn{2}{c|}{Architecture} & \multicolumn{2}{c}{Loss} \\
        \midrule
        $N_{\text{epoch}}$ & 2,000 & $T$ & 40 & $w_\theta$ & 1.0 \\
        $N_{\text{decay,1}}$ & 1,200 & $r_\text{mask}$ & 0.8 & $w_{\Delta \theta}$ & 10.0 \\
        $N_{\text{decay,2}}$ & 1,600 & $D_z$ & 128 & $w_{x}$ & 10.0 \\
        \textit{lr} & 1e-4 & $D_{\text{Tr},z}$ & 256 & $w_{\Delta x}$ & 100.0 \\
         & & $D_{\text{Tr}, \text{ff}}$ & 1024 & $w_J$ & 1.0 \\
         & & $N_{\text{Tr}, \text{layer}}$ & 4 & $w_V$ & 1.0 \\
         & & $N_{\text{Tr}, \text{head}}$ & 8 & $w_{\text{KL}}^{\text{prior}}$ & 0.4 \\
        \bottomrule
    \end{tabular}
    }
    \end{center}
    \vspace{-1em}
    \caption{
        Hyperparameter setup for the motion prior training phase.
        We classify the hyperparameters into three groups, hyperparameters related to the training, architecture, and the loss weights.
    }
    \label{table:supp_hparams_prior}
\end{table}

%% file: tables/supp_hparams_reuse.tex
\begin{table}[t]
    \begin{center}
    \resizebox{0.9\linewidth}{!}{
    \begin{tabular}{lr|lr|lr}
        \toprule
        \multicolumn{2}{c|}{Train} & \multicolumn{2}{c|}{Architecture} & \multicolumn{2}{c}{Loss} \\
        \midrule
        $N_{\text{epoch}}$ & 2,000 & $T$ & 40 & $w_\theta$ & 1.0 \\
        $N_{\text{decay,1}}$ & 1,200 & $r_\text{mask}$ & 0.2 & $w_{\Delta \theta}$ & 10.0 \\
        $N_{\text{decay,2}}$ & 1,600 & $D_z$ & 128 & $w_{x}$ & 10.0 \\
        \textit{lr} & 1e-4 & $D_I'$ & 128 & $w_{\Delta x}$ & 100.0 \\
         & & $D_{\text{Tr},z}$ & 256 & $w_J$ & 1.0 \\
         & & $D_{\text{Tr},\text{ff}}$ & 1024 & $w_V$ & 1.0 \\
         & & $N_{\text{Tr}, \text{layer}}$ & 4 & $w_{\text{KL}}^{\text{reuse}}$ & 0.1 \\
         & & $N_{\text{Tr}, \text{head}}$ & 8 & $w_\beta$ & 0.1 \\
        \bottomrule
    \end{tabular}
    }
    \end{center}
    \vspace{-1em}
    \caption{
        Hyperparameter setup for the motion prior reusing phase.
        We classify the hyperparameters into three groups, hyperparameters related to the training, architecture, and the loss weights.
    }
    \label{table:supp_hparams_reuse}
    \vspace{-1em}
\end{table}